\documentclass[conference]{IEEEtran}
\IEEEoverridecommandlockouts
\UseRawInputEncoding
\usepackage{cite}
\usepackage{amsmath,amssymb,amsfonts}
\usepackage{algorithmic}
\usepackage{graphicx}
\usepackage{textcomp}
\usepackage{subfigure}
\usepackage{tabularx}
\usepackage{booktabs}
\usepackage{threeparttable}
\def\BibTeX{{\rm B\kern-.05em{\sc i\kern-.025em b}\kern-.08em
    T\kern-.1667em\lower.7ex\hbox{E}\kern-.125emX}}
\begin{document}

\title{An Attribute-Enriched Dataset and Auto-Annotated Pipeline for Open Detection\\
\thanks{\IEEEauthorrefmark{2} The three authors contribute equally to this work.}
}
\author{
	\IEEEauthorblockN{
		Pengfei Qi, 
		Yifei Zhang\IEEEauthorrefmark{2}, 
		Wenqiang Li\IEEEauthorrefmark{2}, 
		Youwen Hu\IEEEauthorrefmark{2} 
		and Kunlong Bai}
	\IEEEauthorblockA{China Mobile Research Institute, Beijing, China}
} 

\maketitle

\begin{abstract}
Detecting objects of interest through language often presents challenges, particularly with objects that are uncommon or complex to describe, due to perceptual discrepancies between automated models and human annotators. These challenges highlight the need for comprehensive datasets that go beyond standard object labels by incorporating detailed attribute descriptions. To address this need, we introduce the Objects365-Attr dataset, an extension of the existing Objects365 dataset, distinguished by its attribute annotations. This dataset reduces inconsistencies in object detection by integrating a broad spectrum of attributes, including color, material, state, texture and tone. It contains an extensive collection of 5.6M object-level attribute descriptions, meticulously annotated across 1.4M bounding boxes. Additionally, to validate the dataset's effectiveness, we conduct a rigorous evaluation of YOLO-World at different scales, measuring their detection performance and demonstrating the dataset's contribution to advancing object detection.
\end{abstract}


\section{Introduction}
Object detection, informed by language cues, is a well-researched field, covering open-vocabulary object detection (OVD) {\cite{ghiasi2022scaling,li2022grounded,minderer2022simple}} and referring expression comprehension (REC) {\cite{li2021referring, luo2020multi, liu2023polyformer}. These tasks employ language as an intrinsic representation, facilitating zero-shot or few-shot object detection via textual inputs.
Building on this foundation, recent works{\cite{du2022learning,gu2021open,shi2023edadet,wu2023aligning,zareian2021open}} have concentrated on improving the alignment between visual and language modalities.

Despite notable advancements in the field, existing datasets {\cite{krasin2017openimages, lin2014microsoft,shao2019objects365}} remain limited by their dependence on standardized object vocabularies, which impedes their adaptability to customized text queries. The primary limitations of object vocabulary datasets in OVD and REC are as follows: (1) Lexical Ambiguity: Concise or partial object names can create confusion and diminish the models' ability to differentiate between similar entities. (2) Insufficient Expressiveness: Sole reliance on object names for detection queries may fail to capture the full range of descriptive information. In many real-world contexts, individuals often describe objects based on observable attributes—such as pattern, color, and texture—rather than adhering strictly to formal terminology.
 
\begin{figure}[ht]
\centering
\subfigure[Visualization of Objects365-Attr]{\label{fig1:a}
\includegraphics[width=4.1cm,height=4cm]{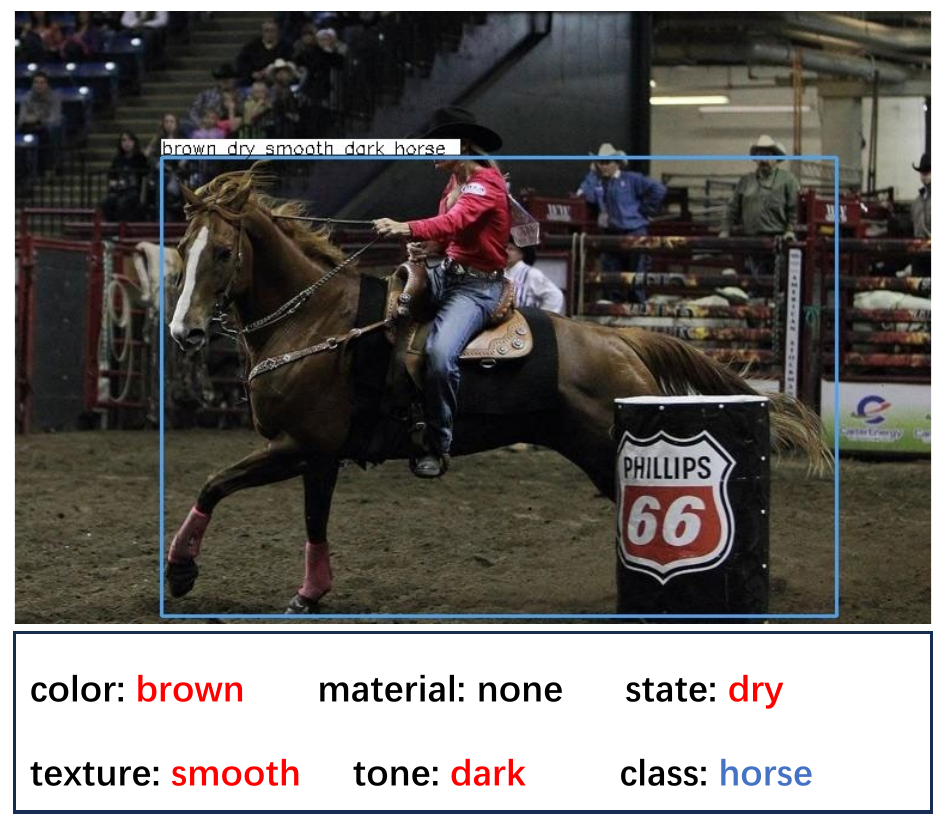}}
\hspace{0.01\linewidth}
\subfigure[Attribute Hierarchy]{\label{fig1:b}
\includegraphics[width=4.2cm,height=4.2cm]{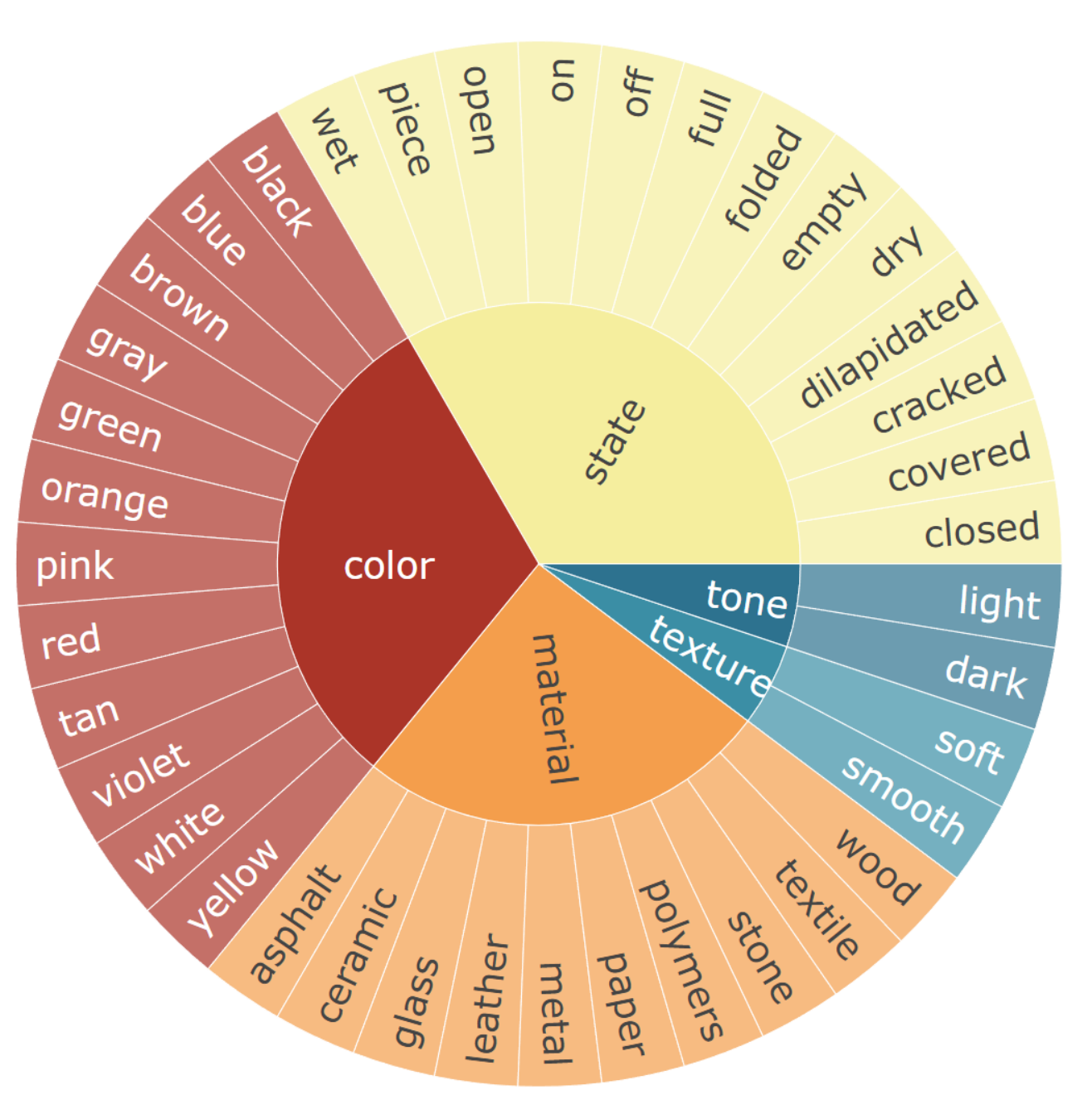}}
\caption{(a) illustrates an example from the Objects365-Attr dataset, generated through the auto-annotated pipeline. The goal is to systematically output all visual attributes for each category represented in the image. (b) shows that the five major categories and their corresponding 39 subcategories within the Objects365-Attr dataset.}
\label{fig1}
\end{figure}

To address these limitations, employing attributes such as color, shape, texture, pattern, and motion as descriptive anchors presents a promising strategy. For instance, as shown in Fig.\ref{fig1}(a), the description ``a brown smooth horse'' often conveys more information than simply using the term ``horse''. 
Compared to vanilla categories, attributes provide three key advantages: (1) Enhanced Contextualization: Attributes can compensate for missing contextual information, thereby offering greater completeness for ambiguous categories. (2) Improved Interpretability: For unfamiliar categories, attributes can be mapped to known ones, facilitating easier interpretation by pre-trained language models. (3) Detailed Representation: Attributes provide a more nuanced description for categories that are otherwise challenging to characterize.

\begin{figure*}[ht]
    \centering
    \includegraphics[width=0.9\linewidth]{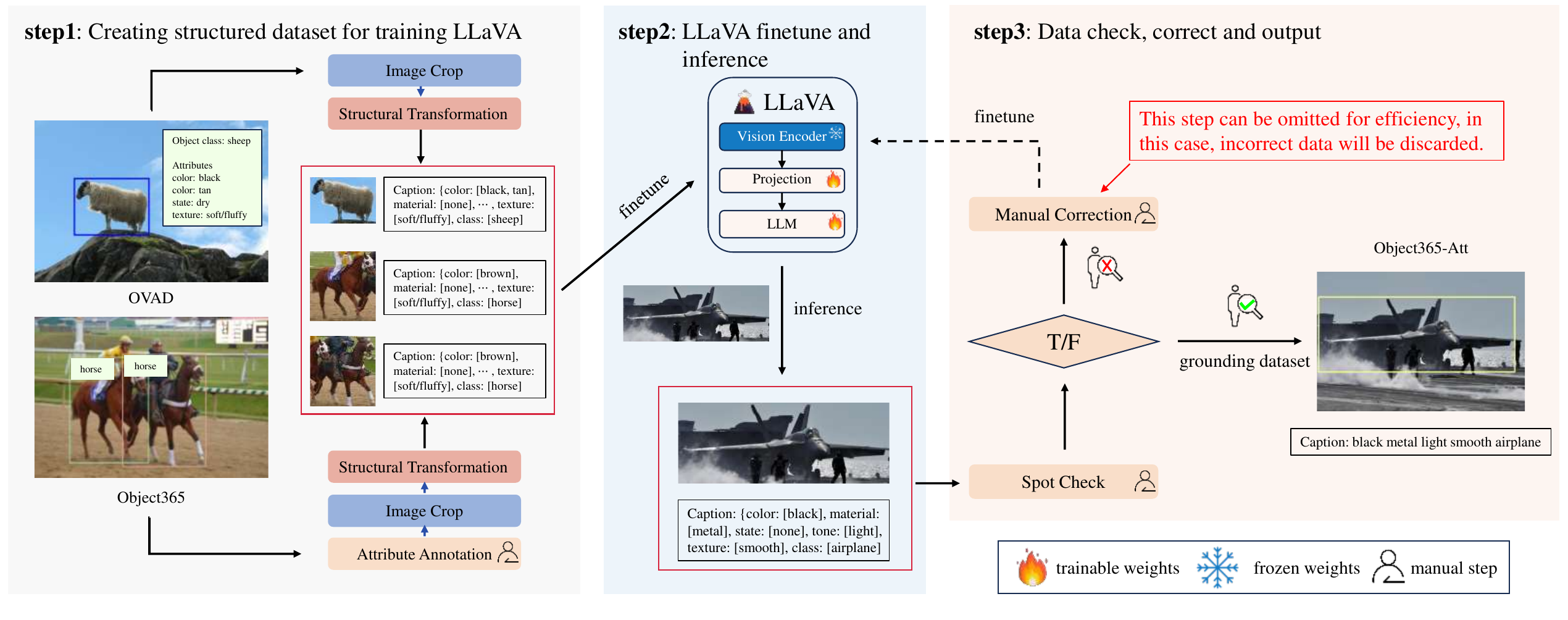}
    \caption{The Auto-Annotated Pipeline. It consists of three key steps, including creating structured dataset for training LLaVA, LLaVA finetune and inference, Data check, correct and output. These steps effectively leverage the image understanding and text generation capabilities of multimodal large models, combining existing attribute datasets with a small amount of human involvement to form our automatic annotation pipeline.}
    \label{Auto-Annotated Pipeline}
\end{figure*}

Inspired by this, our research reorients the paradigm of object detection from mere ``naming'' to utilizing ``descriptive attributes'' by introducing the Objects365-Attr dataset. This dataset not only enables the identification of familiar objects via attribute-based descriptions but also enhances the capacity to articulate the characteristics of unfamiliar objects. The Objects365-Attr is an extension of the Objects365 {\cite{shao2019objects365}} dataset, offering attribute descriptions that span all its categories. To optimize the annotation process, we implement an efficient annotation pipeline designed to streamline annotation tasks while upholding rigorous benchmarking standards. A benchmark is constructed based on descriptive attributes to verify enhanced capabilities for unfamiliar features.

Furthermore, we demonstrate the value of the dataset by evaluating the object-level attribute information learned by YOLO-World {\cite{cheng2024yolo}} methods at different scales. The results indicate that the success of current object detection models is substantially influenced by their capacity to utilize attribute data effectively. 

To sum up, our contributions lies in three folds:
\begin{itemize}
	\item Development of a large-scale attribute description dataset, Objects365-Attr, addressing  shortcomings in existing OVD and REC datasets.
    \item Design of an auto-annotated annotation pipeline that enhances dataset efficiency and accuracy.
	\item Execution of exhaustive experiments and ablation studies that underscore the significance of attribute descriptions in advancing object detection methods.
\end{itemize}

\section{Datasets}

\subsection{Selection of Dataset}\label{AA}

The Objects365 dataset is notable for its extensive scale, encompassing 638K images, broad category coverage with 365 fine-grained classes, and precise annotations, including a total of 10,101K high-quality bounding boxes. It spans 11 major categories, ranging from everyday items to natural and industrial environments, offering more comprehensive content and greater annotation accuracy than datasets such as COCO {\cite{lin2014microsoft}} and OpenImages {\cite{kuznetsova2020open}}. Consequently, we selected Objects365 as the foundational dataset and enhanced it by adding detailed object attribute annotations, resulting in the final Objects365-Attr dataset. The main innovations include:

We selected five key attribute features—color, material, state, texture, and tone—as annotation categories for the Objects365-Attr dataset. These attributes enable the comprehensive identification and description of objects from multiple dimensions. Furthermore, the Objects365-Attr dataset enhances the utilization of visual information by integrating innovative attribute annotations with textual descriptions, effectively linking unique features to specific categories. This approach allows models to leverage these attributes to recall target objects in ambiguous or previously unseen categories. Subsequent experiments have demonstrated that incorporating the Objects365-Attr dataset into the pre-training process substantially improves the performance of OVD and REC.

\subsection{Auto-Annotated Pipeline}
This research employs the Objects365 dataset, enhancing it with an automated process to produce Objects365-Attr. The new dataset integrates categorical data with 39 adjective descriptions, enhancing object semantics. We will now provide a detailed account of the key steps in our automated annotation process in Fig. \ref{Auto-Annotated Pipeline}:

\textbf{Step1: Create structured dataset for training LLaVA.} For fine-tuning the LLaVA model, we prepared a structured dataset comprising two components:

1) OVAD Dataset {\cite{bravo2023open}}: This dataset contains 2,000 images randomly selected from the MS-COCO validation set, covering 14,300 object instances.  Each image is annotated with 117 attributes, meticulously labeled to ensure accuracy and consistency.  From these, 39 common attributes were selected and categorized into five main groups.  We cropped each object instance based on its bounding box, generating cropped images and corresponding prompts and structured outputs.

2) Objects365 Dataset: To diversify our dataset, we randomly sampled 2,000 images from the Objects365 dataset.  Following the same process as the OVAD Dataset, each image instance was annotated with 39 attributes, cropped, and paired with prompts and structured outputs.

\textbf{Step2: LLaVA finetune and inference.} In this step, we fine-tuned the LLaVA-13B {\cite{liu2024visual,liu2024improved}} model using prompts, cropped images, and structured outputs from our initial dataset preparation.  This process enhanced the model's adaptability to specific tasks, thereby improving its performance within the auto-annotation pipeline.  We carefully selected hyperparameters to ensure accurate responses and output generation.  For the remaining images in the Objects365 dataset, we cropped instances based on bounding box annotations and generated corresponding prompts.  During inference, we observed performance limitations when processing low-resolution images, leading us to exclude instances smaller than $100\times100$ pixels.  We then input the filtered prompts and cropped images into the fine-tuned LLaVA model to obtain structured outputs.

\textbf{Step3: Data check, correct and output.} To ensure data quality, we randomly sampled 5\% of the data generated in step 2 for manual inspection.  If the error rate exceeded 2\%, we corrected the errors and merged the revised data with the step 1 data to retrain the LLaVA-13B model.  This inspection and correction process was repeated until all data passed the quality check.  Statistical analysis revealed that some instances, such as those labeled ``person'' lacked attribute information.  Therefore, we filtered out instances without attributes to finalize the dataset.

\subsection{Dataset Statistics}
We proposed Objects365-Attr dataset includes 364 categories from the original Objects365 dataset (excluding the person category) and expands them by adding five major adjective categories, totaling 39 adjective subcategories, as illustrated in the Fig. \ref{fig1:b}. The dataset consists of two parts: a training set and a testing set, for the training set, it comprises 450,651 images and 1,401,491 annotated bounding boxes, with each box containing an average of four adjective dimensions. This enriched annotation approach provides a more detailed and descriptive dataset, enabling models to better understand and detect objects in complex scenarios.

Our goal is to incorporate attribute descriptions into the class query so that the model can more accurately detect the targets we need. During the training phase, considering issues of efficiency and cost, we adopt the method of adjective stacking (e.g. ``a brown dark horse''). However, in actual application scenarios, it is challenging for people to describe an object across so many dimensions, and this method is not efficient enough. People often prefer to provide just one or two adjectives to more accurately detect the target. Starting from this point, we use the same pipeline as for constructing the training set to modify the Object365 validation set (which does not overlap with the training set). The only difference is that each caption contains only one-dimensional adjectives (e.g. ``a brown horse'', ``a dark horse''). To ensure a consistent number of test instances across all dimensions, we randomly sample 2,000 instances from those containing a certain dimension, totaling 10,000 instances to form our test set. The evaluation metric is the accuracy of detection, if the Intersection over Union (IoU) between the detected box and the ground truth is greater than 0.5, it is determined to be a positive sample, and otherwise, it is a negative sample.

\section{Experiments}
\subsection{Implementation Details}

YOLO-World was developed using the MMYOLO{\cite{cheng2024yolo}} and MMDetection{\cite{chen2019mmdetection}} toolboxes. We used the AdamW{\cite{zhuang2022understanding}} optimizer with an initial learning rate of 0.001 and a weight decay of 0.05 to regularize the model. Training was conducted over 100 epochs on 8 NVIDIA A800 GPUs, with a batch size of 256 per epoch.

\subsection{Pre-training Experiments}

\textbf{Selection of Pre-training Datasets.} In this study, we selected datasets focused on detection and grounding tasks for the pre-training of YOLO-World. These include Objects365v1, GQA{\cite{hudson2019gqa}}, Flickr30k{\cite{plummer2015flickr30k}}, and our newly proposed Objects365-Attr.We exclude the images from the COCO dataset in GoldG {\cite{kamath2021mdetr}} (GQA and Flickr30k). The detection datasets provide annotations, such as bounding boxes and category labels or noun phrases, enriching the models with visual and semantic information to enhance feature extraction and object recognition.

\textbf{Zero-shot Evaluation on Our TestSet, LVIS, and Refcoco.} In Tables \ref{testset}, \ref{lvis}, and \ref{refcoco}, we evaluate the impact of incorporating the Objects365-Attr dataset during the pre-training phase of the YOLO-World model using zero-shot testing methods. The inclusion of Objects365-Attr consistently improved the performance of all YOLO-World versions. On our custom test set, pre-training with this dataset led to significant gains in detection accuracy by effectively leveraging attribute information, indicating that in practical applications, enhanced attribute descriptions can yield better results,indicating that in real-world applications, better results can be inferred by increasing attribute descriptions. On the LVIS benchmark, pre-training with Objects365-Attr resulted in an overall improvement of up to 0.6 AP, with a notable increase of 3.1 AP in rare object detection. These findings demonstrate that enriched attribute descriptions in Objects365-Attr substantially enhance OVD task performance. In the RefCOCO {\cite{kazemzadeh2014referitgame}} benchmark, although the improvements were modest across various metrics, they still indicate that enriched attribute descriptions in Objects365-Attr can also enhance REC task performance.

\begin{table}[b]
    \centering
    \caption{Zero-shot Evaluation on our TestSet}
    \resizebox{\linewidth}{!}{ 
    \begin{tabular}{@{}ccccccc@{}}
        \toprule
        Method & Pre-trained Data & Color & Material & State & Texture & Tone \\ 
        \midrule
        YOLO-World-S  & O365,GoldG & 7.5 & 5.5 & 7.5 & 6.6 & 6.9 \\ 
        YOLO-World-S  & O365,GoldG,O365-Attr & 12.6 & 11.3 & 12.1 & 12.1 & 13.0 \\ 
        YOLO-World-L  & O365,GoldG & 7.6 & 5.7 & 7.3 & 6.8 & 7.6 \\ 
        YOLO-World-L  & O365,GoldG,O365-Attr & 13.2 & 11.7 & 13.1 & 12.7 & 13.6 \\ 
        YOLO-World-X  & O365,GoldG & 8.1 & 6.1 & 7.4 & 7.3 & 8.1 \\ 
        YOLO-World-X  & O365,GoldG,O365-Attr & \textbf{13.8} & \textbf{12.3} & \textbf{13.4} & \textbf{13.3} & \textbf{14.2} \\ 
        \bottomrule
    \end{tabular}
    }
    \begin{tablenotes}
    \footnotesize
    \item\textbf{Note:} We evaluate YOLO-World on own TestSet and report Top-1 accuracy for all models. In the case of inference using additional adjectives, training with our Objects365-Attr has an advantage in detecting more correct objects.
    \end{tablenotes}
    \label{testset}
\end{table}

\begin{table}[t]
    \centering
    \caption{Zero-shot Evaluation on LVIS}
    \resizebox{\linewidth}{!}{ 
    \begin{tabular}{@{}cccccc@{}}
        \toprule
        Method  & Pre-trained Data & AP & APr & APc & APf \\ 
        \midrule
        YOLO-World-S  & O365,GoldG &  17.3 & 11.3 & 14.9 & 22.7 \\ 
        YOLO-World-S  & O365,GoldG,O365-Attr & 17.4 & 12.8 & 14.5 & 22.5 \\ 
        YOLO-World-L  & O365,GoldG & 26.0 & 18.6 & 23.0 & \textbf{32.6} \\ 
        YOLO-World-L  & O365,GoldG,CC3M† & 26.1 & 20.6 & 22.6 & 32.3 \\ 
        YOLO-World-L  & O365,GoldG,O365-Attr & \textbf{26.6} & \textbf{21.7} & \textbf{23.3} & 32.4 \\ 
        \bottomrule
    \end{tabular}
    }
    \begin{tablenotes}
    \footnotesize
    \item\textbf{Note:} We evaluate YOLO-World on LVIS val in a zero-shot manner and report AP fo all models. CC3M†{\cite{sharma2018conceptual}} represents a pseudo-labeled dataset proposed by YOLO-World.
    \end{tablenotes}
    \label{lvis}
\end{table}

\begin{table}[t]
    \centering
    \caption{Zero-shot Evaluation on RefCOCO}
    \resizebox{\linewidth}{!}{ 
    \begin{tabular}{@{}ccccc@{}}
        \toprule
        Method & Pre-trained Data & RefCOCO & RefCOCO+ & RefCOCOg \\ 
         &  & val testA testB & val testA testB & test\\
        \midrule
        YOLO-World-S  & O365,GoldG & 26.9 33.9 18.0 & 33.8 18.3 27.4 & 29.8\\
        YOLO-World-S  & O365,GoldG,O365-Attr & 26.9 \textbf{35.0} 18.1 & \textbf{35.2} 17.8 27.2 & 29.2\\ 
        YOLO-World-L  & O365,GoldG &  27.2 33.9 18.8 & 34.8 19.9 27.5  & 30.5 \\ 
        YOLO-World-L  & O365,GoldG,O365-Attr & \textbf{27.9} 34.4 \textbf{19.7} & 34.5 \textbf{20.5} \textbf{27.7} & \textbf{30.6}\\ 
        \bottomrule
    \end{tabular}
    }
    \begin{tablenotes}
    \footnotesize
    \item\textbf{Note:} We evaluate YOLO-World on RefCOCO, RefCOCO+, RefCOCOg in a zero-shot manner and report Top-1 fo all models.
    \end{tablenotes}
    \label{testset}
    \label{refcoco}
\end{table}
\subsection{Ablation Experiments}

Table \ref{ablation} presents a series of stepwise ablation experiments conducted on the pre-training datasets using YOLO-World-L, demonstrating that the inclusion of the Objects365-Attr dataset significantly improves object detection performance.

Pre-training with the original Objects365 dataset (Num. 1) followed by the addition of the Objects365-Attr data (Num. 2) led to a 1.5$\%$ increase in the AP metric. Similarly, when using the original Objects365 and GoldG datasets as the pre-training baseline (Num. 3), the integration of the Objects365-Attr dataset (Num. 5) resulted in a substantial 3.1$\%$ improvement in APr performance.

To ensure that these improvements were not merely due to an increase in the volume of pre-training data, we controlled for the data quantity. We standardized the base pre-training data to include the original datasets and then separately added the Objects365-Attr data and a dataset with an equivalent number of images labeled only by category names. Under these conditions, the model utilizing the Objects365-Attr data (Num. 5) outperformed the model trained with only category-labeled data (Num. 4) by 1.1$\%$ in AP and achieved a notable 2.3$\%$ improvement in APr.

\begin{table}[t]
\centering  
\caption{Ablation experiments}
\resizebox{\linewidth}{!}{ 
\begin{tabular}{@{}ccccccc@{}}  
\toprule
Num.& \multicolumn{2}{c}{Pretrain Data} & AP & APr & APc & APf  \\ 
  & Base data & Add Data & & &  &\\
\midrule
1 & O365 & $\backslash$ &17.3 & 7.7 & 14.3 & 24.8   \\  
2 & O365  &O365-Attr & \textbf{18.8}& \textbf{11.0} & \textbf{15.8} & \textbf{25.6} \\  
\midrule
3 & O365+GoldG  &$\backslash$ &26.0 & 18.6 & 23.0 & \textbf{32.6} \\ 
4 & O365+GoldG  &O365-Attr† & 25.5 & 19.4 & 22.2 & 31.8\\  
5 & O365+GoldG  &O365-Attr & \textbf{26.6} & \textbf{21.7} & \textbf{23.3} & 32.4  \\  

\bottomrule
\end{tabular} 
}
\begin{tablenotes}
\footnotesize
\item\textbf{Note:} O365-Attr† refers to a dataset that contains the same number of images as in objects365-Attr but with annotations limited to only the category names.
\end{tablenotes}
\label{ablation} 
\end{table}

\subsection{Qualitative Experiments}

Experiments results of pre-trained YOLO-World-L in two settings: 1) Fig. \ref{Qualitative experiments} (a), (b) show zero-shot inference results with LVIS rare categories, compared to original YOLO-World model, (a) our model (b) improves the ability to detect rare categories. 2) Fig. \ref{Qualitative experiments} (c), (d) show inference results with Attribute description plus class name, compared to original YOLO-World model (c) use attributes to add category as a reminder, our model (d) can better detect objects. 3) Fig. \ref{Qualitative experiments} (e), (f) respectively show that compared with outputting category names with original YOLO-World model, using noun phrases with added attributes for inference with model after pre-training on our dataset can more accurately recall the targets. 

\begin{figure}[t]
\resizebox{\linewidth}{!}{ 
\centering
\begin{tabular}{cccc} 
    (a)   &  \includegraphics[width=5.5cm,height=3.8cm]{vis-imgs/1.jpg} & \includegraphics[width=5.5cm,height=3.8cm]{vis-imgs/2.jpg} & \includegraphics[width=5.5cm,height=3.8cm]{vis-imgs/3.jpg}\\
    (b)  &  \includegraphics[width=5.5cm,height=3.8cm]{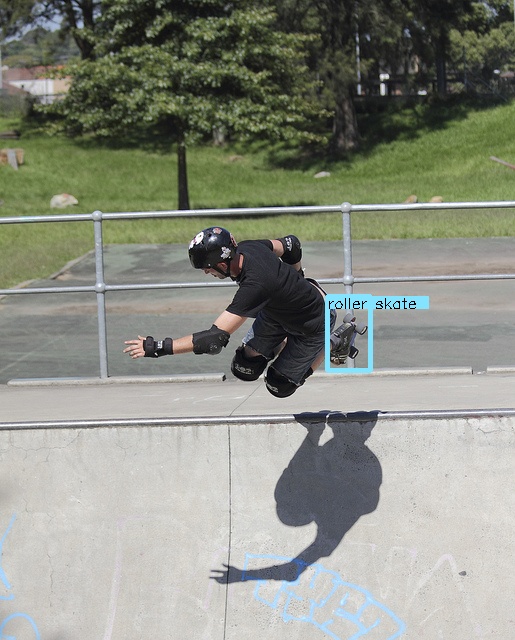} & \includegraphics[width=5.5cm,height=3.8cm]{vis-imgs/ovad2.jpg} & \includegraphics[width=5.5cm,height=3.8cm]{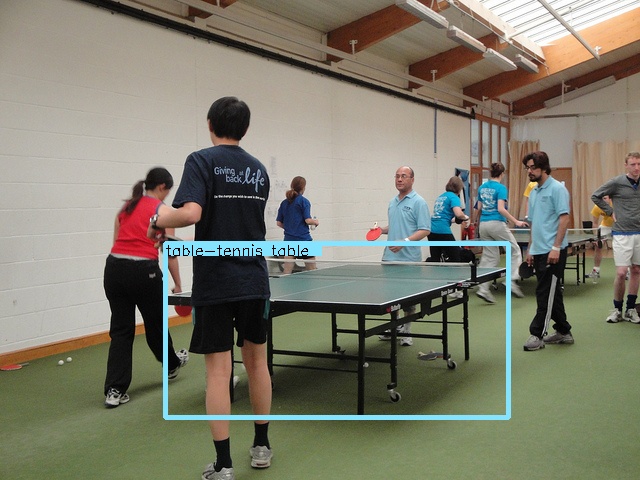} \\
    & roller skate & stepladder & table-tennis table \\
    (c)   &  \includegraphics[width=5.5cm,height=3.8cm]{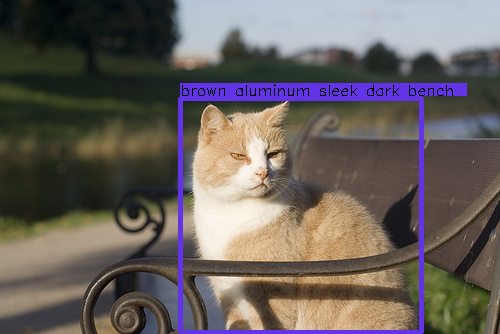} & \includegraphics[width=5.5cm,height=3.8cm]{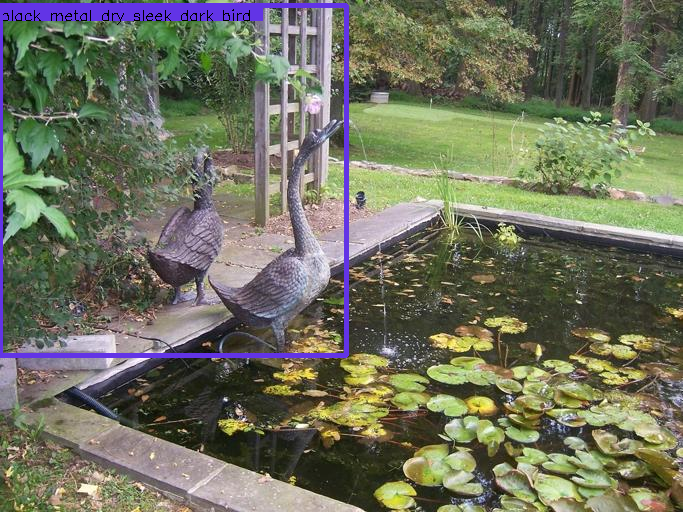} & \includegraphics[width=5.5cm,height=3.8cm]{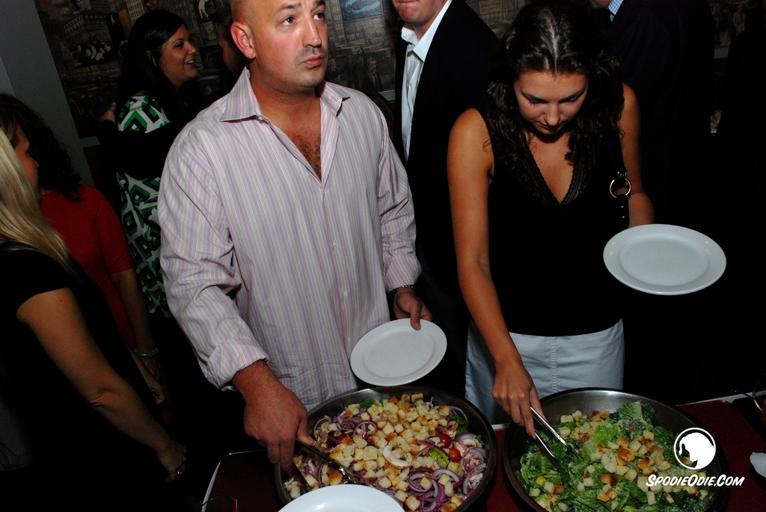} \\
    (d)  &  \includegraphics[width=5.5cm,height=3.8cm]{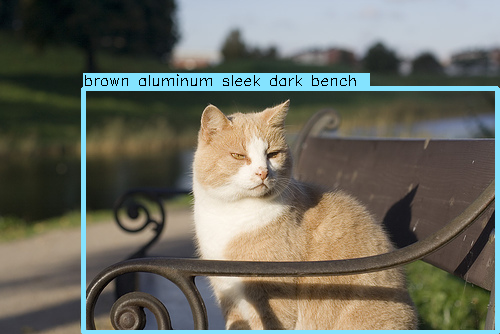} & \includegraphics[width=5.5cm,height=3.8cm]{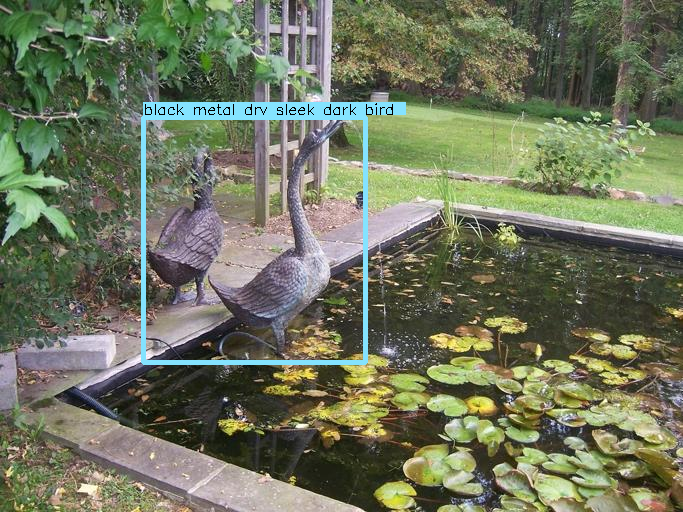} & \includegraphics[width=5.5cm,height=3.8cm]{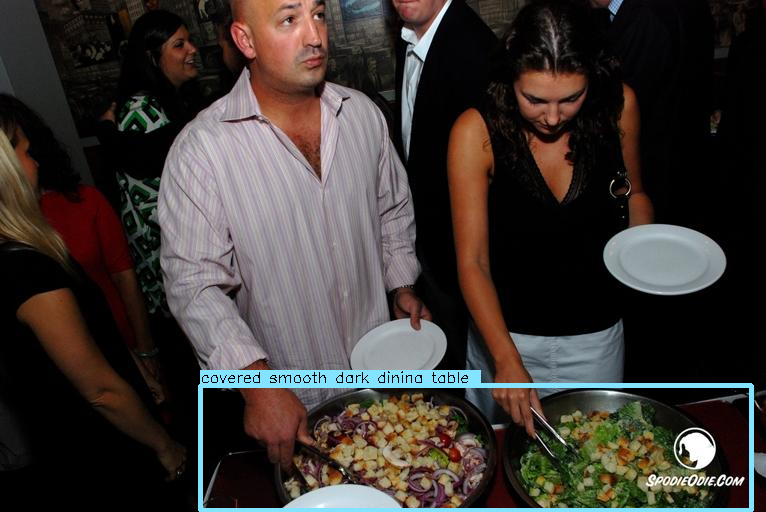} \\
    & brown aluminum sleek dark bench & sleek metal dry dark bird & covered smooth dark dining table \\
    (e)   &  \includegraphics[width=5.5cm,height=3.8cm]{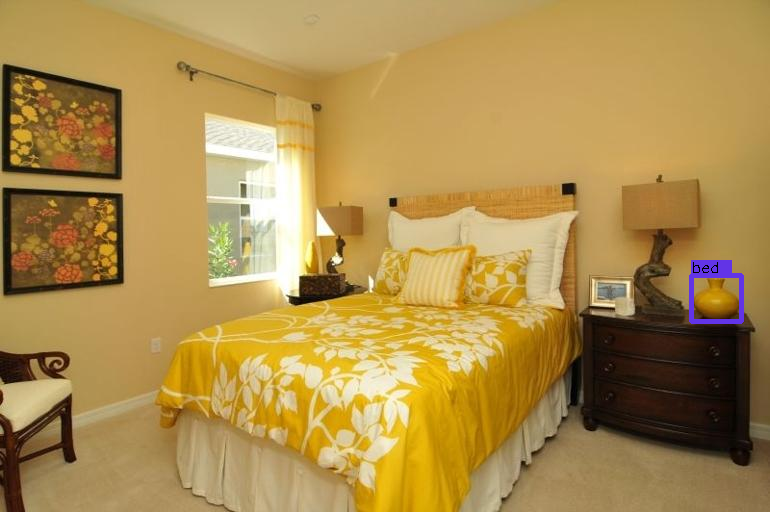} & \includegraphics[width=5.5cm,height=3.8cm]{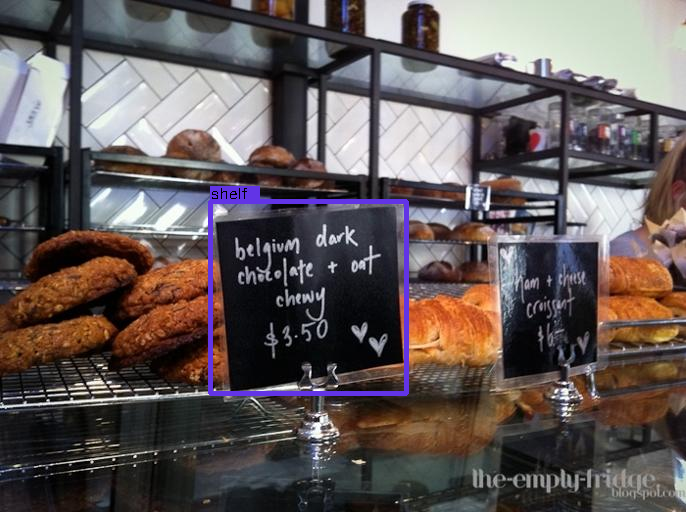} & \includegraphics[width=5.5cm,height=3.8cm]{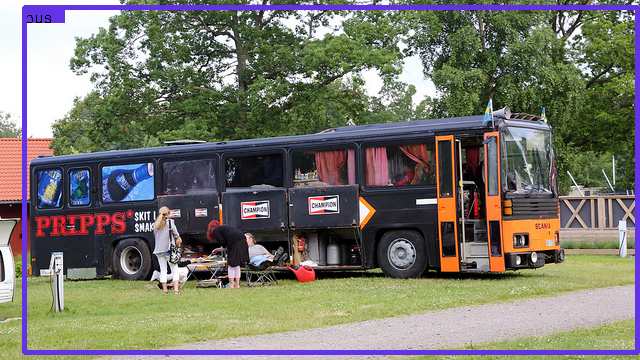} \\
    & bed & shelf & bus \\
    (f)  &  \includegraphics[width=5.5cm,height=3.8cm]{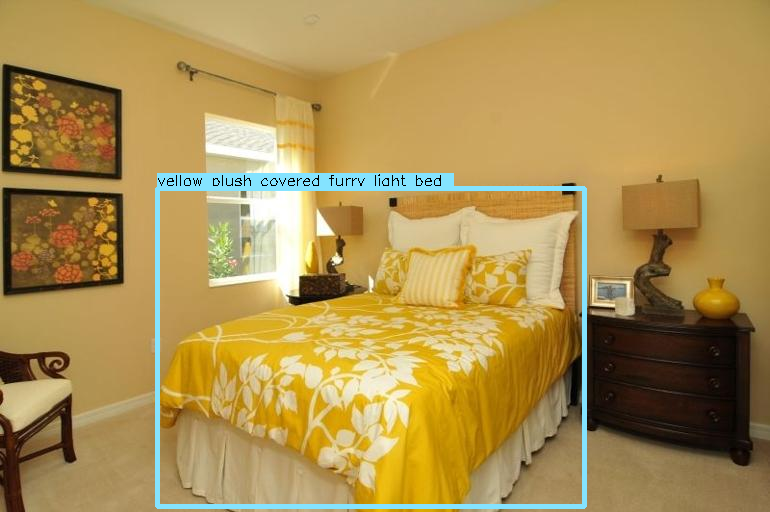} & \includegraphics[width=5.5cm,height=3.8cm]{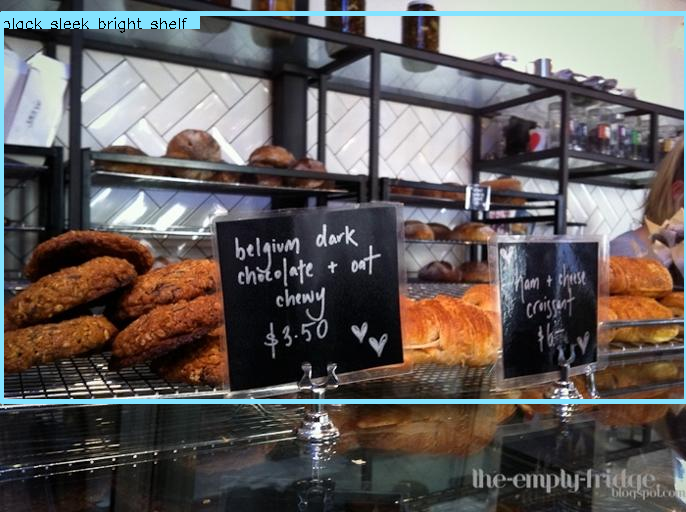} & \includegraphics[width=5.5cm,height=3.8cm]{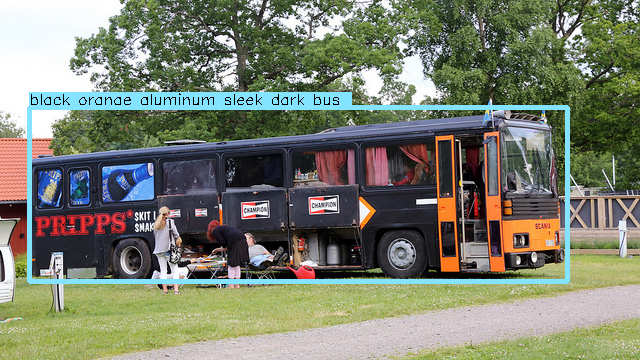} \\
    & yellow plush coverd furry light bed & black sleek bright shelf & black orange aluminum sleek dark bus \\
\end{tabular}
 }
 \caption{ displays the inference results using the YOLO-World-L weights,shows the inference results after additional pre-training with the Objects365-Attr dataset. (a), (b) for the visualization on the LVIS dataset, focusing exclusively on the rare categories. (c), (d) for the visualization of a class name with attributes. (e), (f) are visualizations of different weights resulting from inputting different prompts.}
 \label{Qualitative experiments}
\end{figure}

\section*{Conclusion}

In this paper, we explore the use of attribute descriptions for open-vocabulary detection and provide detailed attribute annotations for an existing dataset. Building on this, we introduce the Object365-Attr dataset, which annotates objects using flexible attribute expressions. By evaluating models based on fundamental capabilities, Object365-Attr serves as a comprehensive and in-depth training resource for reliable and thorough studies. We believe that this dataset and our findings will advance the understanding and development of open-vocabulary detection methods, thereby facilitating future research in this field.

\bibliographystyle{IEEEtran}
\bibliography{IEEEabrv,icassp}

\end{document}